\documentclass[letterpaper, 10 pt, conference]{ieeeconf}

\IEEEoverridecommandlockouts
\usepackage{cite}
\usepackage{epsfig}
\usepackage{mathptmx}
\usepackage{times}
\usepackage{amsmath,amssymb,amsfonts}
\usepackage{algorithmic}
\usepackage{graphicx}
\usepackage{textcomp}
\usepackage{xcolor}
\usepackage{subcaption}
\usepackage{glossaries}
\usepackage{hyperref}

\def\BibTeX{{\rm B\kern-.05em{\sc i\kern-.025em b}\kern-.08em
T\kern-.1667em\lower.7ex\hbox{E}\kern-.125emX}}

\begin{document}

\newacronym{batman}{B.A.T.M.A.N}{Better Approach To Mobile Ad-hoc Networking }
\newacronym{dds}{DDS}{Data Distribution Service}
\newacronym{dymu}{DyMu}{Dynamic-Multi-Layered Path Planning}
\newacronym{esa}{ESA}{European Space Agency}
\newacronym{esric}{ESRIC}{European Space Resources Innovation Centre}
\newacronym{fps}{fps}{frames per second}
\newacronym{gui}{GUI}{graphical user interface}
\newacronym{imu}{IMU}{Inertial Measurement Unit}
\newacronym{isru}{ISRU}{In-Situ Resources Utilisation}
\newacronym{lidar}{LiDAR}{Light Detecting And Ranging}
\newacronym{lro}{LRO}{Lunar Reconnaissance Orbiter}
\newacronym{mrs}{MRS}{Multi-Robot Systems}
\newacronym{realms}{REALMS}{Resilient Exploration And Lunar Mapping System}
\newacronym{realms2}{REALMS2}{Resilient Exploration And Lunar Mapping System 2}
\newacronym{rgbd}{RGB-D}{RGB-Depth}
\newacronym{ros}{ROS}{Robot Operating System}
\newacronym{ros2}{ROS 2}{Robot Operating System version 2}
\newacronym{slam}{SLAM}{Simultaneous Localisation And Mapping}
\newacronym{snt}{SnT}{Centre for Security, Reliability and Trust}
\newacronym{roi}{ROI}{region of interest}
\newacronym{rtabmap}{RTAB-Map}{Real-Time Appearance Based Mapping}
\newacronym{viper}{VIPER}{Volatiles Investigating Polar Exploration Rover}
\newacronym{vslam}{vSLAM}{Visual Simultaneous Localisation And Mapping}
\newacronym{v&v}{V\&V}{Verification and Validation}
\newacronym{hwmp}{HWMP}{Hybrid Wireless Mesh Protocol}
\newacronym{sas}{SAS}{Space Applications Services}
\newacronym{manet}{MANET}{Mobile Ad-Hoc Networks}
\newacronym{per}{PER}{Packet Error Rate}
\newacronym{cots}{COTS}{Commercial-of-the-Shelf}
\newacronym{ransac}{RANSAC}{Random sample consensus}
\newacronym{rls}{RLS}{Recursive Least Squares}
\newacronym{darpa}{DARPA}{Defense Advanced Research Projects Agency}
\newacronym{rmse}{RMSE}{Root Mean Square Error}
\newacronym{lstm}{2D LSTM}{2D convolutional Long Short Term Memory (LSTM)}
\newacronym{sgp}{SGP}{Sparse Gaussian processes}

\newif\ifanonymized
\anonymizedfalse      

\ifanonymized
\title{RoughSense: Lightweight Terrain-Induced Rover Vibration Prediction Using Point Clouds and IMU Feedback
\thanks{This research was funded in whole, or in part, by Anonymous research institution, grant reference [Anonymous]. For the purpose of open access, and in fulfilment of the obligations arising from the grant agreement, the author has applied a Creative Commons Attribution 4.0 International (CC BY 4.0) license to any Author Accepted Manuscript version arising from this submission.}
}
\author{Anonymous 1$^{1}$, Anonymous 2$^{2}$, Anonymous 3$^{1}$%
\thanks{$^{1}$ From Anonymous center, Anonymous}%
\thanks{$^{2}$ From Anonymous center, Anonymous}%
}
\hypersetup{pdfauthor={}}

\else
\title{RoughSense: Lightweight Terrain-Induced Rover Vibration Prediction Using Point Clouds and IMU Feedback
\thanks{This research was funded in whole, or in part, by the Luxembourg National Research Fund (FNR), grant reference [CAESAR-XR/Gabriel/17679211]. For the purpose of open access, and in fulfilment of the obligations arising from the grant agreement, the author has applied a Creative Commons Attribution 4.0 International (CC BY 4.0) license to any Author Accepted Manuscript version arising from this submission.}
}
\author{Gabriel Manuel Garcia$^{1}$, St{\'e}phanie Aravecchia$^{2}$, Miguel Angel Olivares-Mendez$^{1}$%
\thanks{$^{1}$ From University of Luxembourg, Luxembourg}%
\thanks{$^{2}$ From IRL Georgia Tech-CNRS, Metz, France}%
}
\fi

\maketitle

\begin{abstract}
Autonomous navigation in space requires reliable terrain assessment for safe operations, especially in underground environments with limited communication, computing resources, and power budget. This paper presents a lightweight method for real-time vibration-aware traversability mapping using a \gls{lidar} point cloud and \gls{imu} measurements. An initial vibration proxy is estimated from terrain geometry by applying \gls{ransac} to local point-cloud patches produced by a \gls{slam} algorithm. In parallel, the \gls{imu} provides local observations of the vibration experienced by the rover during traversal. The point-cloud-based prediction is then corrected online using Recursive Least Squares, allowing the system to adapt the geometric estimate to the measured rover response. The approach is evaluated in a lunar analogue environment, an outdoor field, and an underground mine.
\end{abstract}


\glsresetall
\section{Introduction}

In recent years, the new space movement has changed the space sector, with an increasing number of companies and space agencies targeting distant and unexplored environments \cite{space_growth}. Although space exploration is becoming more feasible, efficient exploration still requires a high degree of autonomy because remote planetary environments induce communication delays, limited bandwidth, and reduced opportunities for direct teleoperation.

There is also a growing interest in underground environments, as highlighted by the \gls{darpa} Subterranean Challenge \cite{b_slam_darpa}, and in planetary lava tubes on the Moon or Mars \cite{interest_lunar_under}. These environments can provide protection from radiation, micrometeorites, and extreme temperature variations, but they also introduce poor illumination, communication loss, uneven terrain, and limited localisation. In such conditions, a rover cannot rely continuously on an operator and must use its own autonomous navigation system.

\begin{figure} [htp]
    \centering
    \includegraphics[width=1\columnwidth]{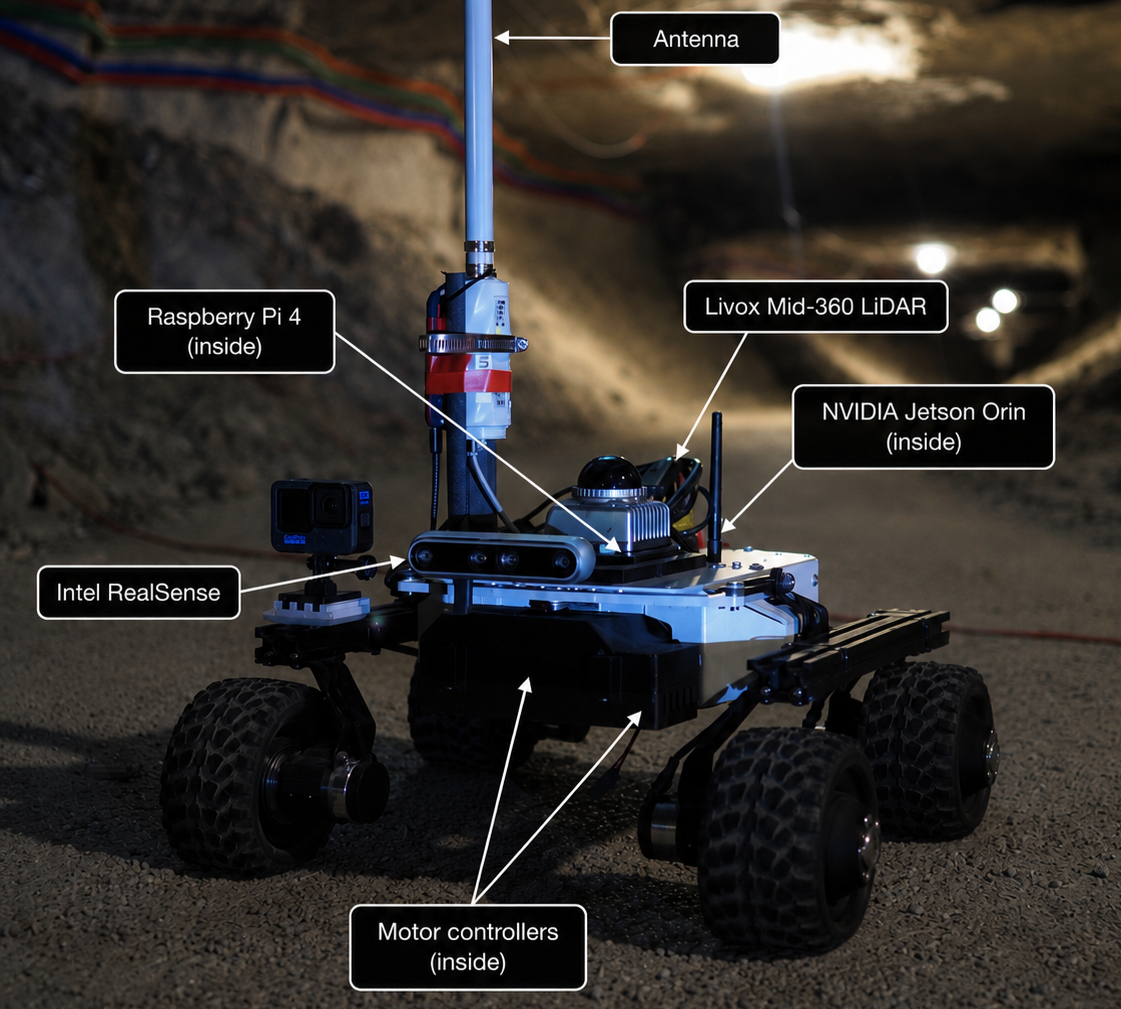}
    \caption{Leo Rover in the Walferdange mine.}
    \label{fig:leo_rover}
\end{figure}

Autonomous navigation in these environments requires more than obstacle avoidance. A collision-free path may still be inefficient or unsafe if the terrain induces strong vibrations, high mechanical stress, or unstable motion. Therefore, path planning should rely on terrain cost maps that describe not only whether a region is traversable, but also how the terrain may affect the rover during traversal \cite{b_occupancy_grid_example}. In this work, we focus on terrain-induced robot vibration prediction as a lightweight traversability cue for rough and unstructured environments.

This paper presents a real-time approach that estimates a vibration cost map by combining exteroceptive and proprioceptive sensing. A point cloud generated by a \gls{lidar} is used to estimate an initial vibration proxy from terrain geometry. Local planes are fitted to the point cloud using the \gls{ransac} algorithm, and the residual distribution is used to quantify the expected terrain-induced vibration. Since this geometric estimate cannot fully capture the robot-specific response, \gls{imu} measurements are used as local observations of the vibration actually experienced by the rover. The predicted vibration values are then corrected online using the lightweight \gls{rls} algorithm.

The main contributions of this paper are:

\begin{itemize}
    \item A lightweight methodology for predicting terrain-induced rover vibration using both exteroceptive and proprioceptive sensors.
    \item An online adaptive correction method based on \gls{rls}, which reduces the mismatch between point-cloud-based prediction and \gls{imu}-based observation and propagates the correction over a cost map.
    \item Experimental validation in three representative environments: a lunar analogue facility, a lake shore, and an underground mine.
\end{itemize}

The proposed approach is evaluated in the Lunalab, at Symphony Lake's shores, and in the Walferdange mine, demonstrating online adaptation of geometric terrain predictions using local vibration observations under real-time space robotics constraints.

\section{Related work}
This section reviews related work on rover vibration prediction for terrain traversability analysis. Existing approaches can be broadly divided according to the sensing modality used to estimate the terrain-induced response of the robot. Proprioceptive approaches rely on onboard measurements of the robot response during traversal, while exteroceptive approaches use external perception sensors to analyse the terrain before contact. Finally, this section discusses online adaptation methods, which are relevant for correcting exteroceptive predictions using local proprioceptive observations under the computational constraints of space robotics.

\subsection{Proprioceptive approach}
Proprioceptive sensing has been widely used to infer terrain properties from the physical interaction between a mobile robot and the ground. Brooks and Iagnemma showed that wheel--terrain interaction induces characteristic vibrations in a planetary rover structure, which can be used for online terrain classification from accelerometer measurements \cite{b_proprio_1}. Weiss et al. extended this idea using support vector machines and compared several acceleration representations, including raw, frequency-domain, and statistical features \cite{b_proprio_2}. Bai et al. further investigated three-dimensional acceleration signals with frequency-domain features and a neural-network classifier, highlighting that vibration measurements can capture terrain properties that are not directly observable from geometry or appearance alone \cite{b_proprio_3}. These works demonstrate that proprioceptive signals provide direct information about the robot--terrain interaction. However, they mainly formulate the problem as discrete terrain classification after contact, whereas the present work focuses on predicting a continuous robot vibration response before traversal.

\subsection{Exteroceptive approach}
The main limitation of purely proprioceptive methods is that they can only analyse terrain already traversed by the rover. Exteroceptive sensors, such as \gls{lidar} and depth cameras, can instead perceive the terrain before contact. Point clouds are commonly used for traversability analysis and terrain assessment \cite{b_std_2}, \cite{b_survey_1}, \cite{b_survey_2}, \cite{b_std_3}. A common geometric approach consists of fitting a local plane to the terrain and analysing the point-to-plane residuals, often using the \gls{rmse}, following \cite{b_ransac} and \cite{b_std_intro_2}. Related methods use similar geometric indicators, such as the roughness index introduced in \cite{b_std_1}, based on the standard deviation of elevation adjusted by ground clearance. In this work, such geometric roughness descriptors are interpreted as lightweight exteroceptive proxies for the vibration that the rover may experience during traversal.

Deep learning methods can also process point clouds or RGB images to estimate terrain traversability \cite{b_nn}, \cite{b_nn_2}. However, image-based approaches are less suited to underground environments because of poor lighting and the energy cost of additional illumination. More generally, learning-based methods may be limited for space robotics by their computational requirements, dependence on training data, and reduced interpretability \cite{b_nn_not_working}. Therefore, this work focuses on lightweight geometric descriptors rather than deep neural-network-based terrain analysis.

\subsection{Online adaptation}
Purely exteroceptive terrain descriptors do not fully capture the robot-specific vibration response, which also depends on morphology, wheels, suspension, speed, and terrain mechanical properties. Online adaptation is therefore useful to correct geometric predictions using proprioceptive observations acquired during traversal. Waibel et al. compared a \gls{lidar}-based terrain-cost method using \gls{ransac} with an LSTM-based method predicting future \gls{imu} measurements along possible paths \cite{b_2d_convolutional}. Their results show the benefit of learning-based adaptation, since learned models can reduce the need for manually tuned thresholds and better account for complex terrain--robot interactions. However, such approaches remain computationally demanding and may require significant training data, which limits their suitability for space-constrained systems.

To address this limitation, the present work uses a lightweight online linear regression method based on \gls{rls} \cite{b_rls_intro}. \gls{rls} has been used for recursive model identification and correction, including in combination with Kalman filtering \cite{b_rls_kalman}, and can efficiently integrate new observations in real time \cite{b_rls_1}. This work uses \gls{rls} to adapt the relationship between point-cloud-based terrain roughness and measured \gls{imu} vibration. The objective is not to replace geometric terrain analysis, but to correct its output online using the actual vibration experienced by the rover. This provides a computationally efficient framework for robot vibration prediction under space robotics constraints.

Overall, the literature highlights a gap between exteroceptive traversability estimation, which predicts before contact but remains an indirect proxy for the robot response, and proprioceptive vibration analysis, which measures the actual interaction only after traversal. This work addresses this gap by combining point-cloud-based prediction with lightweight online correction from \gls{imu} feedback.

\begin{figure*} [htp]
    \centering
    \includegraphics[width=1.6\columnwidth]{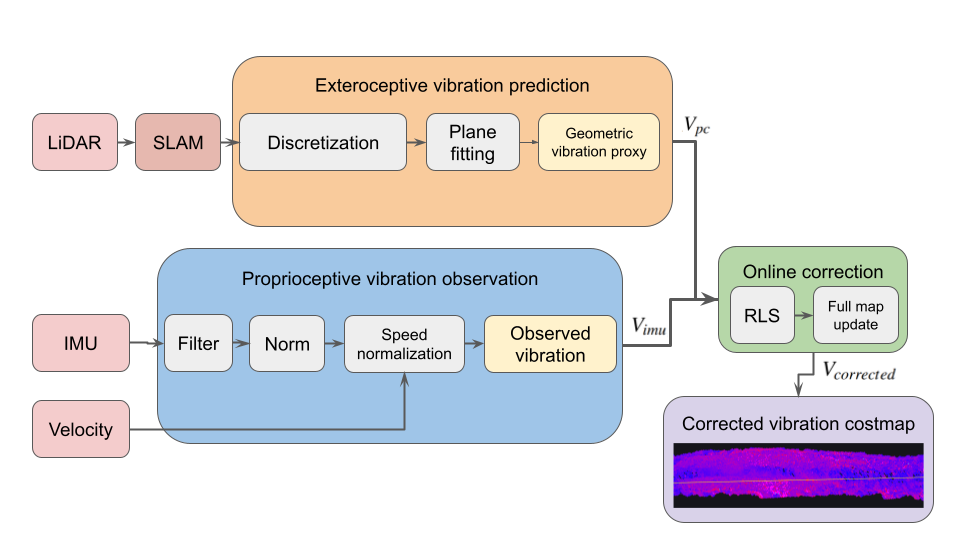}
    \caption{Overview of the proposed vibration prediction framework. A point-cloud-based exteroceptive prediction $V_{pc}$ and an \gls{imu}-based proprioceptive observation $V_{imu}$ are combined through an online \gls{rls} correction to produce the corrected vibration cost map $V_{corrected}$.}
    \label{fig:overview}
\end{figure*}

\section{Methodology}
\label{sec:method}
Fig.~\ref{fig:overview} shows an overview of the proposed architecture. In this framework, a point cloud is used as an exteroceptive input to estimate a terrain-induced vibration proxy before traversal, while \gls{imu} data provides local observations of the vibration experienced by the rover after contact. Once both the predicted and observed vibration scores have been computed, an online correction is applied to adapt the point-cloud prediction to the measured robot response. This correction relies on the \gls{rls} algorithm and produces a corrected vibration cost map.

\subsection{Exteroceptive vibration prediction}
The initial prediction is computed from a \gls{lidar} point cloud, although the method applies to any sensor producing a point cloud. \gls{lidar} is well suited to underground and space-relevant environments, where limited lighting makes vision-based perception less practical. Its power consumption can also be comparable to stereo vision systems: Mastcam-Z requires around $20\,\mathrm{W}$ for imaging \cite{lidar_justif_1}, while an industry-grade \gls{lidar} can produce point clouds with around $10\,\mathrm{W}$ \cite{lidar_justif_2}.

Once the point cloud, accumulated in the map frame from the \gls{slam} algorithm, is gathered, it is discretised into a local grid representing the surroundings of the robot. The \gls{ransac} algorithm \cite{b_ransac} is then applied to each cell of the grid to fit a local plane, which is used as a geometric reference for the terrain surface.

After fitting the plane, the distance between each point and the plane is calculated. A distribution of points close to the plane indicates a locally flat terrain, while a scattered distribution indicates a more irregular terrain. Following \cite{b_ransac}, we consider the residuals of the planes fitted to patches of the point cloud. Then, following \cite{b_std_2}, we define the point-cloud vibration proxy $\hat{V}_{pc}$ as the \gls{rmse} of the residuals, with $N$ the number of points used to fit the plane and $r_i$ the residual:
\begin{equation}
\hat{V}_{pc} = \sqrt{\frac{\sum_{i=0}^{N} (r_i)^2}{N}}
\label{equ:pc_vibration}
\end{equation}

This value is not a direct measurement of vibration, but a geometric proxy for the vibration that the rover may experience when traversing the corresponding terrain region. Once $\hat{V}_{pc}$ is calculated, it is normalised using a fixed \textit{min--max} calibration chosen before the experiments. The minimum value is set to $0$, while the maximum value is manually selected from a barely traversable area. The normalised predicted vibration score is denoted $V_{pc}$, such that $V_{pc} \in [0,1]$, where higher values correspond to stronger expected vibration and more difficult terrain. Since the descriptor is computed from residuals around a locally fitted plane, it remains independent of the terrain slope.

\subsection{Identification of rover-induced vibration frequencies}
\label{sec:method_res_freq}
The raw \gls{imu} signal contains both terrain-induced vibrations and vibrations generated by the rover itself. To identify these rover-induced components, two experiments were performed with the same velocity command $v$: a \textit{stationary} experiment, where the rover was elevated on foam and its wheels rotated without ground contact, and a \textit{moving} experiment, where the rover drove over small rocks. Frequency peaks appearing in both spectra were attributed to the rover and removed using a notch stop-band filter before computing the observed vibration score.

\subsection{Proprioceptive vibration observation}
Even with a precise exteroceptive sensor, prediction errors remain because the point cloud may not capture all terrain properties that influence the robot response. For example, terrain compliance, loose material, hidden obstacles, or robot-specific dynamics can affect the measured vibration without being fully represented by the point-cloud geometry. To obtain a local observation of the vibration actually experienced by the rover, we use the linear acceleration measured by the \gls{imu}:
\begin{equation}
S_{imu} =
\begin{bmatrix}
L_x\
L_y\
L_z
\end{bmatrix}
\label{equ:signal_imu}
\end{equation}

Because the \gls{imu} signal contains rover-induced vibration frequencies identified in Sec.~\ref{sec:method_res_freq}, these components are removed using a notch stop-band filter before further processing:
\begin{equation}
S^{f}_{imu} = f*{notch}(S_{imu})
\label{equ:signal_filtered}
\end{equation}

We estimate the instantaneous vibration magnitude $\hat{N}_{imu}$ as the $L_2$ norm of the filtered linear acceleration, as indicated in Eq.~\ref{equ:norm_imu}. This produces a single signal representing the overall translational excitation of the rover body:
\begin{equation}
\hat{N}_{imu} = |S^{f}_{imu}|_2
\label{equ:norm_imu}
\end{equation}

The vibration magnitude is then normalised by the rover speed $v$, as indicated in Eq.~\ref{equ:norm_speed}. Since the rover velocity can be zero when the platform is not moving, a small threshold $\epsilon$ is used to avoid division by zero:
\begin{equation}
N_{imu} = \frac{\hat{N}_{imu}}{\max(v,\epsilon)}
\label{equ:norm_speed}
\end{equation}

Next, the observed vibration score $\hat{V}_{imu}$ is computed as the \gls{rmse} of $N_{imu}$ over a sliding window of size $k=100$ \gls{imu} samples, with $\mu$ the mean value of the signal over the window:
\begin{equation}
\hat{V}_{imu} =  \sqrt{\frac{\sum*{i=0}^{k} (N_{imu,i}-\mu)^2}{k}}
\label{equ:v_imu}
\end{equation}

Finally, a fixed \textit{min--max} normalisation is applied to $\hat{V}_{imu}$. As for $V_{pc}$, the minimum value is set to $0$, while the maximum value is manually selected from a barely traversable area before the experiments. The normalised observed vibration score is denoted $V_{imu}$, such that $V_{imu} \in [0,1]$, where higher values correspond to stronger measured vibration.

\subsection{Online vibration correction}
The point-cloud vibration proxy $V_{pc}$ provides an estimate before traversal, while the \gls{imu}-based score $V_{imu}$ provides a local observation of the actual rover response after contact. Since both scores are normalised using fixed calibration values and share the same convention, the point-cloud prediction can be corrected online using local proprioceptive observations.

The \gls{rls} algorithm is selected for this task because it is lightweight, recursive, and capable of integrating new observations in real time. Following \cite{b_rls_intro}, the \gls{rls} algorithm is executed in three main steps. The first step computes the gain vector, as shown in Eq.~\ref{equ:rls_kalman_gain}. The forgetting factor is set to $\lambda=0.995$:
\begin{equation}
k_{n} = \frac{P_{n-1}x_n}{\lambda+x_n^TP_{n-1}x_n}
\label{equ:rls_kalman_gain}
\end{equation}
with $x_n = [V_{pc}, 1]^T$ the feature vector used in the correction. In the implementation, a denominator floor is used to avoid numerical instability.

The parameter vector is then updated according to Eq.~\ref{equ:rls_parameter_update}:
\begin{equation}
\theta_n = \theta_{n-1} + k_n (V_{imu} - x_n^T\theta_{n-1})
\label{equ:rls_parameter_update}
\end{equation}
with:
\begin{equation}
\theta_n =
\begin{bmatrix}
\alpha\
\beta
\end{bmatrix}
\end{equation}

In Eq.~\ref{equ:rls_parameter_update}, $\theta_n$ represents the estimated correction parameters that map the exteroceptive prediction $V_{pc}$ to the observed proprioceptive vibration $V_{imu}$. To improve robustness, the implementation includes residual gating, parameter clipping, and a small covariance inflation term when numerical issues are detected.

After calculating the updated parameters, the covariance matrix is updated using Eq.~\ref{equ:rls_cov_update}. This covariance matrix is then used for the next parameter update:
\begin{equation}
P_n = \frac{1}{\lambda}(P_{n-1}-k_nx_n^TP_{n-1})
\label{equ:rls_cov_update}
\end{equation}

The corrected vibration score is finally computed as:
\begin{equation}
V_{corrected}(i,j) = \alpha V_{pc}(i,j) + \beta
\label{equ:corrected_vibration}
\end{equation}

The estimated parameters are applied to the full vibration cost map, allowing local \gls{imu} observations to adapt the exteroceptive prediction over the mapped terrain. The resulting corrected score combines the predictive capability of exteroceptive terrain perception with the local accuracy of proprioceptive vibration measurements.

\section{Experimental results}

\begin{figure*}[h!]
    \centering
    \begin{subfigure}[b]{0.6\columnwidth}
        \centering
        \includegraphics[width=\textwidth]{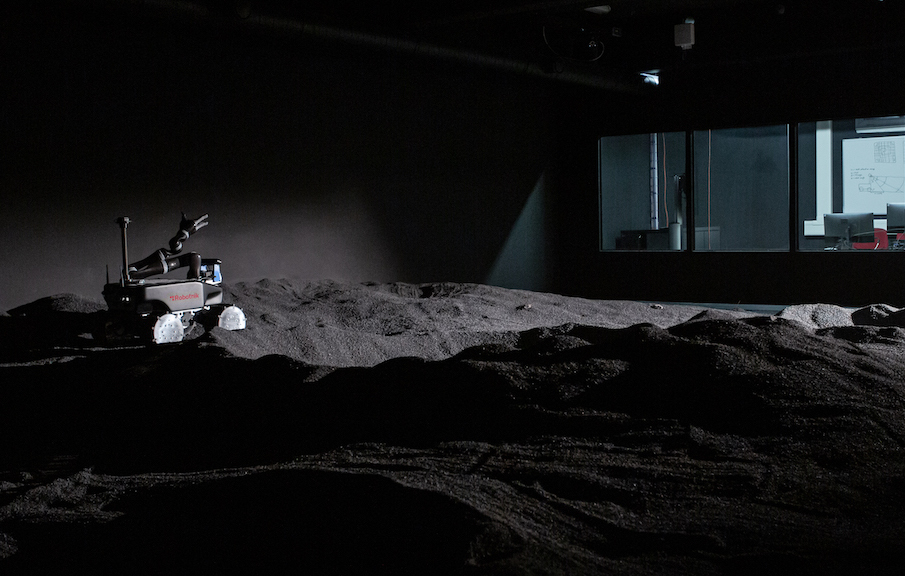}
        \caption{Lunalab}
        \label{fig:lunalab}
    \end{subfigure}
        \begin{subfigure}[b]{0.72\columnwidth}
        \centering
        \includegraphics[width=\textwidth]{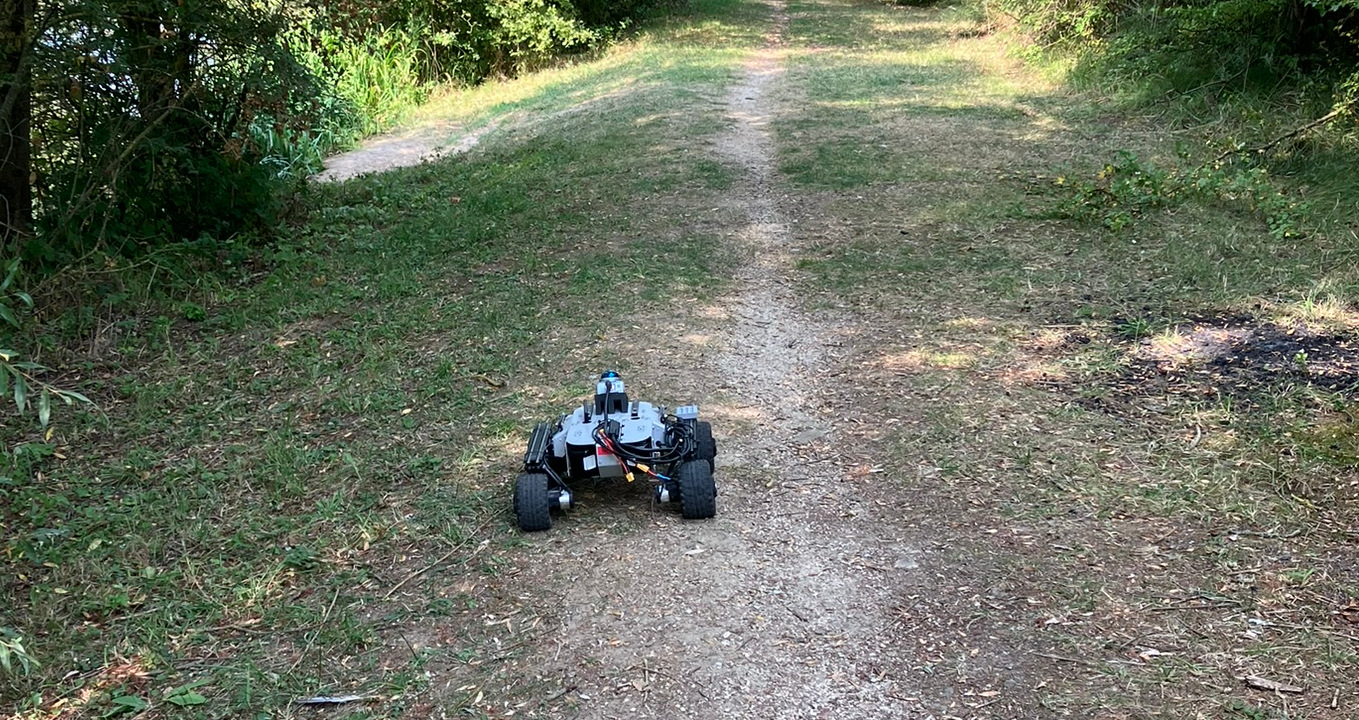}
        \caption{Symphony Lake's shores}
        \label{fig:georgia}
    \end{subfigure}
    \begin{subfigure}[b]{0.69\columnwidth}
        \centering
        \includegraphics[width=\textwidth]{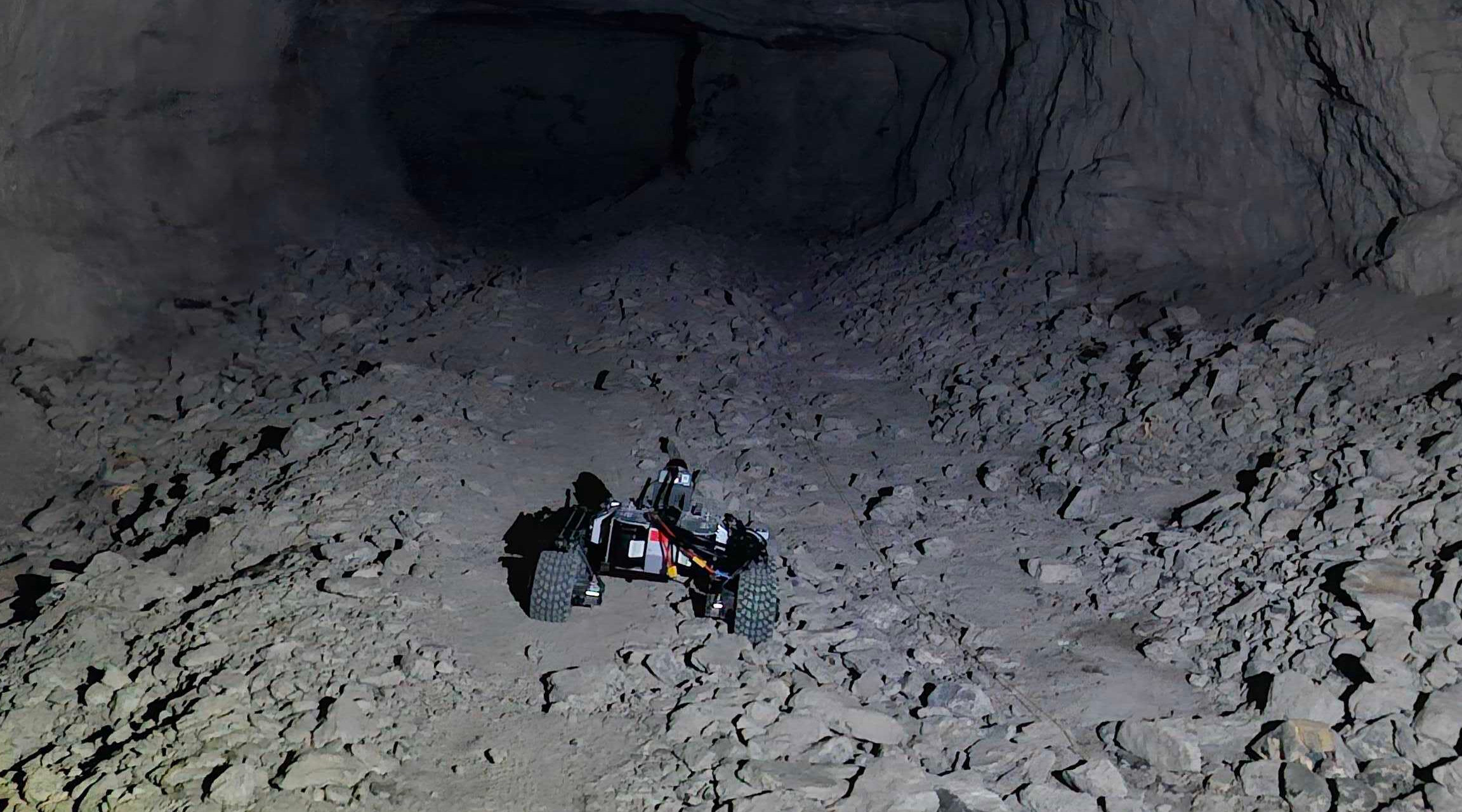}
        \caption{Walferdange mine}
        \label{fig:mine}
    \end{subfigure}
    \caption{Experimental environments. Fig.~(a) shows the lunar analogue facility at the University of Luxembourg, the Lunalab. Fig.~(b) shows the shores of Symphony Lake. Fig.~(c) shows a tunnel view of the experiments conducted in the Walferdange mine.}
    \label{fig:environments}
\end{figure*}

\subsection{Experimental setup and environments}
The robotic platform is based on the \textit{Leo Rover} \cite{b_leo} and was equipped with an Intel RealSense D455, a Livox MID-360 \gls{lidar} \cite{b_livox}, and an \gls{imu}. The \gls{lidar} operates at $10\,\mathrm{Hz}$ and provides $200,000$ points/s with a range from $0.1\,\mathrm{m}$ to $40\,\mathrm{m}$, while the \gls{imu} operates at $200\,\mathrm{Hz}$. Three constant velocity commands were used: $0.1\,\mathrm{m/s}$, $0.3\,\mathrm{m/s}$, and $0.5\,\mathrm{m/s}$. All calculations were performed on one CPU core of an \textit{NVIDIA Jetson Orin}, without GPU acceleration.

The \gls{slam} algorithm \textit{RTAB-Map} provides the aggregated point cloud and rover localisation. The vibration prediction algorithm is triggered when a localisation update is received and the rover reaches a new grid cell.

Experiments were conducted in the three environments shown in Fig.~\ref{fig:environments}: Lunalab, a lunar analogue facility at the University of Luxembourg \cite{b_lunalab}; Symphony Lake's shores, an outdoor field environment with rocky terrain; and the Walferdange mine \cite{b_walferdange_mine}, a long underground environment with rough terrain. Together, these environments evaluate the method under controlled, outdoor, and underground conditions.

\subsection{Rover-induced vibration frequencies}
\label{sec:res_freq}
\begin{figure}[htbp]
    \centering
    \begin{subfigure}[b]{0.49\textwidth}
        \centering
        \includegraphics[width=\textwidth]{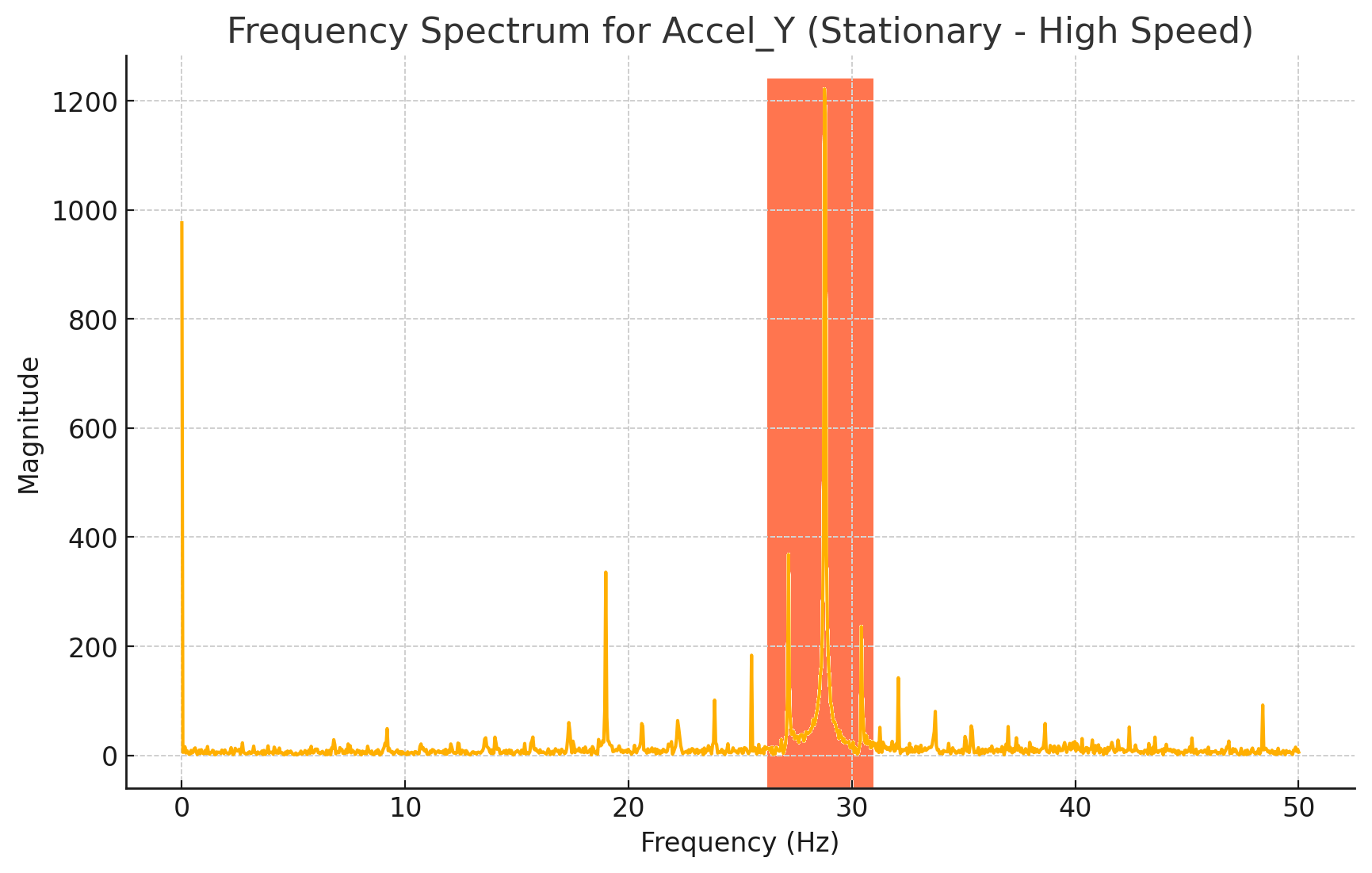}
        \caption{Stationary experiment spectrum}
        \label{fig:spectrum_stationary}
    \end{subfigure}
    \begin{subfigure}[b]{0.49\textwidth}
        \centering
        \includegraphics[width=\textwidth]{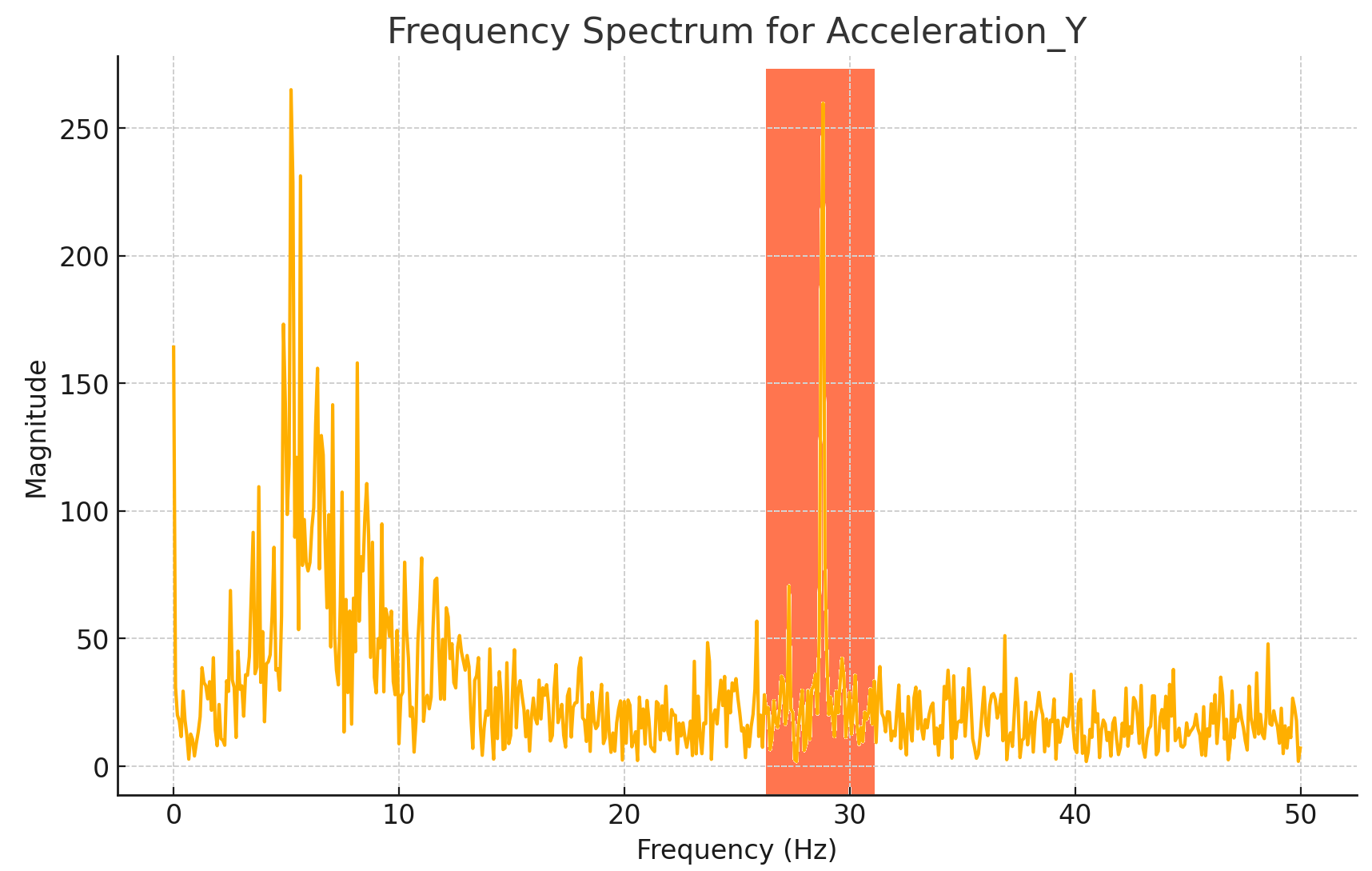}
        \caption{Moving experiment spectrum}
        \label{fig:spectrum_moving}
    \end{subfigure}
    \caption{Frequency spectra of the rover \gls{imu} signal. Fig.~(a) shows the signal spectrum when the rover wheels rotate without terrain contact. Fig.~(b) shows the signal spectrum when the rover drives over small rocks.}
    \label{fig:resonating_freq}
\end{figure}

The frequency analysis described in Sec.~\ref{sec:method_res_freq} was used to identify vibration components generated by the rover itself. Fig.~\ref{fig:resonating_freq} shows the spectra obtained during the \textit{stationary} and \textit{moving} experiments.

The spectrum of the \textit{stationary} experiment shows a dominant peak around $27.5,\mathrm{Hz}$, indicating a vibration component generated by the rover rather than by terrain contact. The same peak is visible in the \textit{moving} experiment, confirming that this component is also present when the rover drives over rocks. To focus the vibration score on terrain-induced motion, this frequency band is removed from the raw \gls{imu} signal using the notch stop-band filter defined in Eq.~\ref{equ:signal_filtered}.

\subsection{Vibration prediction error analysis}
\label{sec:results}


After running the experiments, a post-processing analysis was performed to evaluate the vibration prediction error. For each prediction shift $s$, the point-cloud prediction $V_{pc}$ and the corrected prediction $V_{corrected}$ were evaluated at a future cell located $s$ grid cells ahead of the rover. Once this cell was traversed, both predictions were compared with the observed \gls{imu}-based vibration score $V_{imu}$. The shift $s$ was then converted into a prediction distance using the grid resolution.

The prediction error is evaluated as the absolute difference between each estimate and the observed vibration score:
\begin{equation}
e_{pc} = \lvert V_{pc} - V_{imu} \rvert ,
\label{equ:error_pc}
\end{equation}
\begin{equation}
e_{corrected} = \lvert V_{corrected} - V_{imu} \rvert .
\label{equ:error_corrected}
\end{equation}
The quantities $e_{pc}$ and $e_{corrected}$ are evaluation metrics rather than physical vibration scores. They measure the absolute error between the observed \gls{imu}-based vibration score and the point-cloud-only or corrected prediction, respectively.



\begin{figure*}[htbp]
\centering
\includegraphics[width=\textwidth]{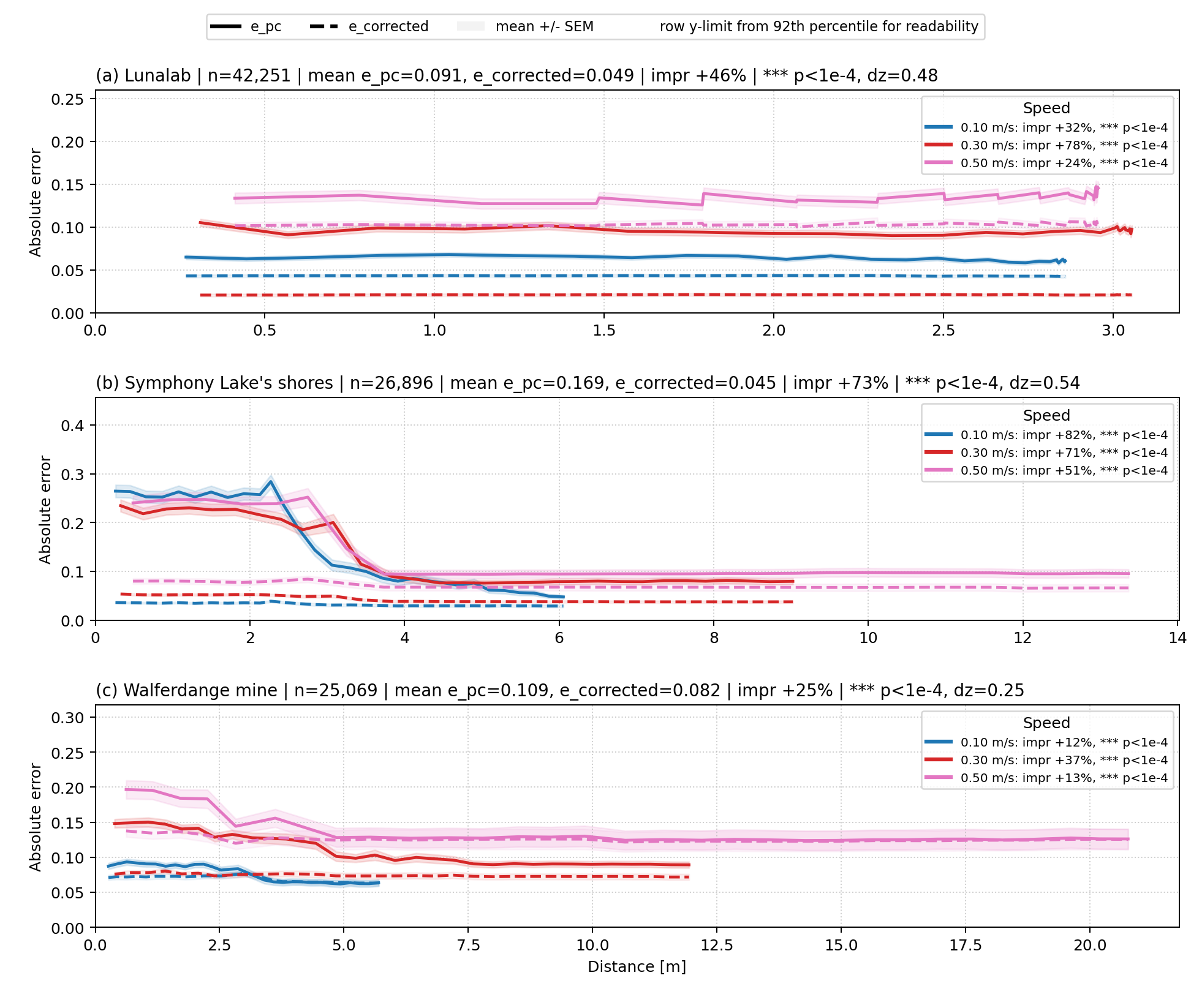}
\caption{Mean absolute vibration prediction error as a function of prediction distance. Solid lines show the point-cloud-only error $e_{pc}$, dashed lines show the corrected error $e_{corrected}$, and shaded regions indicate the standard error of the mean.}
\label{fig:vibration_trend}
\end{figure*}

Fig.~\ref{fig:vibration_trend} shows the evolution of the mean prediction error with respect to prediction distance. Across all environments, the corrected prediction error $e_{corrected}$ remains lower than the point-cloud-only prediction error $e_{pc}$. The overall mean error is reduced from $0.091$ to $0.049$ in Lunalab, from $0.169$ to $0.045$ at Symphony Lake's shores, and from $0.109$ to $0.082$ in the Walferdange mine. This corresponds to an average improvement of $46\%$, $73\%$, and $25\%$, respectively. The correction is therefore most effective in the outdoor rocky environment, while the improvement remains more limited in the mine, where the terrain is more irregular and the prediction range is longer.

\begin{figure*}[htbp]
\centering
\includegraphics[width=\textwidth]{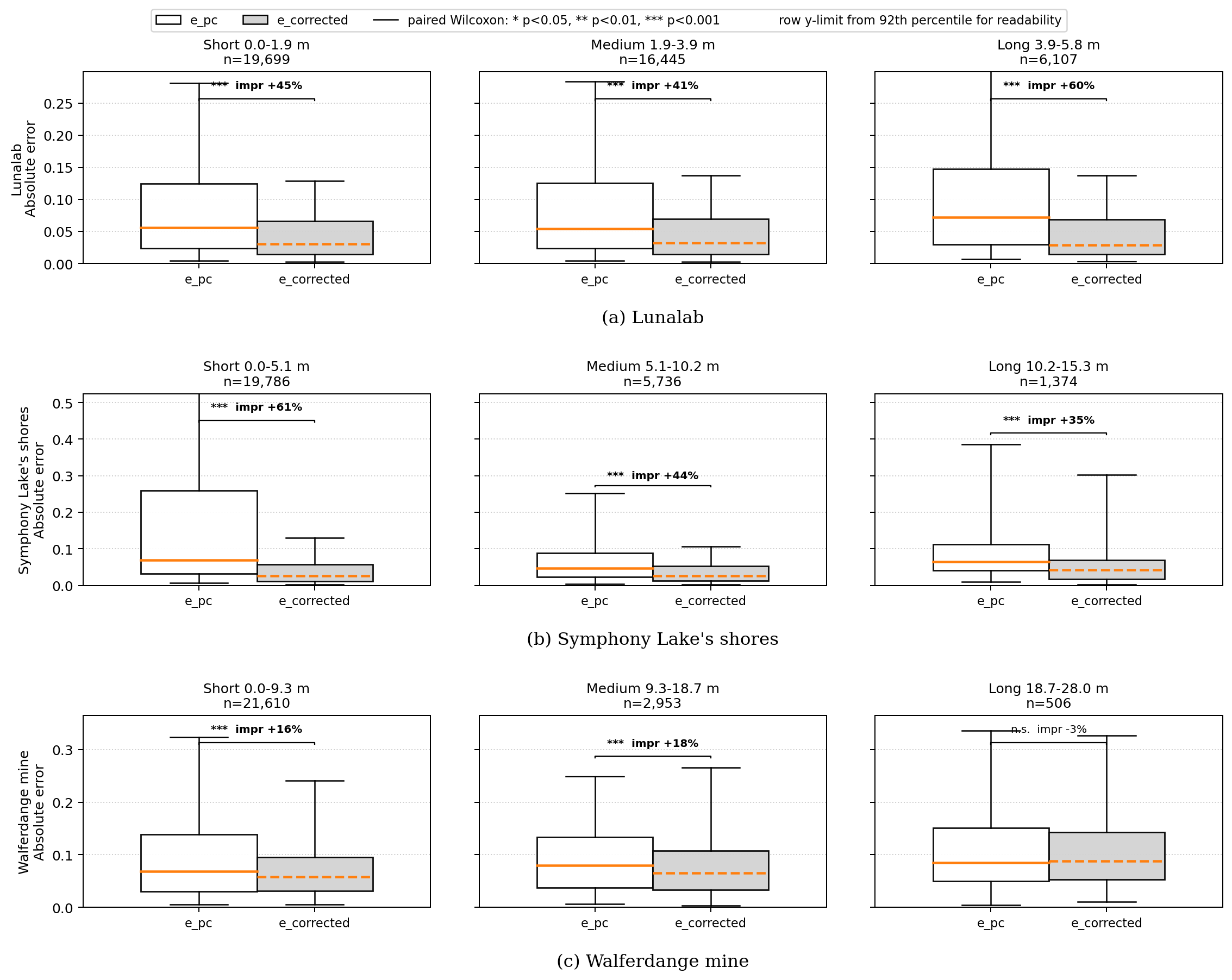}
\caption{Distribution of the absolute vibration prediction error over short-, medium-, and long-range predictions. Each pair compares $e_{pc}$ and $e_{corrected}$; significance is evaluated with a paired Wilcoxon signed-rank test.}
\label{fig:vibration_results}
\end{figure*}

Fig.~\ref{fig:vibration_results} shows the distribution of the prediction errors over short-, medium-, and long-range distance bands. The corrected prediction generally reduces both the median error and the dispersion compared with the point-cloud-only prediction. The improvement is statistically significant in all distance bands except for the long-range prediction in the Walferdange mine, where the correction does not improve the median error. This result is consistent with the reduced number of samples at long distance and the lower point-cloud density available for distant cells. The stabilisation of the error at longer distances is discussed in the limitations section.

\subsection{Computation time}
The \gls{rls} algorithm was selected for its lightweight computational properties. To validate this choice, the computation time of each critical step in the vibration prediction pipeline was measured and is summarised in Table~\ref{tab:time}. The results show that all processing steps are performed within a few milliseconds on one CPU core of the \textit{NVIDIA Jetson Orin}, without GPU acceleration.

\begin{table}[h!]
    \begin{center}
    \caption{Computation time, mean and standard deviation.}
    \label{tab:time}
        \begin{tabular}{ l r l}
        Calculation block          & Time($\mu$s)    &             \\
        \hline
        Vibration prediction       & 42 394 &$\pm$ 12 277       \\
        Vibration observation      & 1 &$\pm$ 0.32              \\
        Vibration correction       & 234 &$\pm$ 34.71           \\
        Publish map (ROS2 header)  & 1 801 &$\pm$ 288.41        \\
        \hline
        Total                      & 44 431&                   \\
        \end{tabular}
    \end{center}
\end{table}

Table~\ref{tab:time} shows that the online correction step requires only $234,\mu\mathrm{s}$ on average, corresponding to approximately $0.53\%$ of the total computation time. The total processing time of approximately $44,\mathrm{ms}$ is compatible with the localisation-triggered update rate used in the experiments. This confirms that the \gls{rls}-based correction adds only a small computational overhead compared with the point-cloud processing stage.


\section{Limitations}
As discussed in Sec.~\ref{sec:results}, the proposed method is sensitive to the grid-cell resolution. Cells that are too small may contain too few point-cloud samples to reliably estimate the local terrain geometry, while cells that are too large may aggregate terrain features at a scale that is not relevant to the rover response. The cell size must therefore be selected according to both the rover footprint and the point-cloud density.
A second limitation comes from the dependence on \gls{slam} and point-cloud density. Localisation inaccuracies can affect the consistency of the vibration cost map, while the \gls{lidar} point density decreases at longer prediction distances. As a result, some distant cells may not contain enough points to compute a reliable exteroceptive vibration prediction. This behaviour is visible in Fig.~\ref{fig:vibration_trend}, where the prediction error tends to stabilise at longer distances, and in Fig.~\ref{fig:vibration_results}, where the long-range prediction in the Walferdange mine shows limited correction improvement.
The proposed method also depends on fixed calibration thresholds used to normalise the point-cloud vibration proxy and the \gls{imu}-based vibration observation. These thresholds were selected from barely traversable areas before the experiments. Although this provides a consistent score convention across datasets, the calibration may need to be adjusted for rovers with different morphology, wheel design, suspension, or operational speed.
Finally, the proposed correction relies on a linear relationship between the point-cloud-based vibration proxy and the \gls{imu}-based vibration observation. While this choice enables lightweight real-time computation, it may not capture all nonlinear effects caused by terrain compliance, wheel--terrain interaction, rover speed, or suspension dynamics.

\glsresetall
\section{Conclusion}
This work introduced a lightweight method for terrain-induced robot vibration prediction using both exteroceptive and proprioceptive sensing. A point-cloud-based geometric descriptor was used to estimate an initial vibration cost before traversal, while local \gls{imu} measurements provided observations of the vibration actually experienced by the rover. The \gls{rls} algorithm was then used to adapt this prediction online and produce a corrected vibration cost map for traversability analysis.

The experimental results show that the proposed correction reduces the prediction error and error dispersion compared with the point-cloud-only prediction, while adding only a small computational overhead. The method was validated in three representative environments: a lunar analogue facility, a lake shore, and an underground mine.

Future work will focus on extending the framework with additional terrain characteristics, such as elevation, slope, obstacle information, and terrain discontinuities, while preserving the low computational cost of the current approach. Further work will also investigate improved vibration models and additional field experiments in space-relevant analogue environments.
The project can be found in this repository: \url{https://github.com/Gabryss/RoughSense/tree/ISPARO2026}


\vspace{12pt}

\end{document}